\documentclass[twoside,leqno,twocolumn]{article}

\usepackage[letterpaper]{geometry}

\usepackage{ltexpprt}
\usepackage[normalem]{ulem}
\usepackage{times}
\usepackage{xcolor}
\usepackage{soul}

\usepackage{amsmath,amssymb,amsfonts}
\usepackage{textcomp}
\usepackage{epsfig,epstopdf}
\usepackage{subfigure}
\usepackage{float}
\usepackage{graphicx}
\usepackage{makecell}
\usepackage{comment}
\usepackage{appendix}
\usepackage[ruled]{algorithm}
\usepackage{algorithmic}
\usepackage[english]{babel}
\usepackage{tabularx}
\usepackage{rotating,multirow}
\usepackage{booktabs}
\usepackage{array}

\begin{document}
\title{Attention-Aware Answers of the Crowd}

\author{Jingzheng Tu$^\dagger$, Guoxian Yu$^\dagger$\thanks{Corresponding Author: Guoxian Yu
(gxyu@swu.edu.cn). This paper was accepted by SDM'2020.}, Jun Wang\thanks{Southwest University,\{tujinzheng, gxyu, kingjun\}@swu.edu.cn.},
	Carlotta Domeniconi\thanks{George Mason University, carlotta@cs.gmu.edu.},
	Xiangliang Zhang\thanks{King Abdullah University of Science and Technology, xiangliang.zhang@kaust.edu.sa}}

\date{}

\maketitle

\fancyfoot[R]{\scriptsize{Copyright \textcopyright\ 20XX by SIAM\\
Unauthorized reproduction of this article is prohibited}}

\begin{abstract}
Crowdsourcing is a relatively  economic and efficient solution to collect annotations from the crowd through online platforms. Answers collected from workers with different expertise may be noisy and unreliable, and the quality of annotated data needs to be further maintained. Various solutions have been attempted to obtain high-quality annotations. However, they all assume that workers' label quality is stable over time (always at the same level whenever they conduct the tasks).
In practice, workers' attention level changes over time,  and the ignorance of which  can affect the reliability of the annotations. In this paper, we focus on a \emph{novel and realistic} crowdsourcing scenario involving \emph{attention-aware annotations}. We propose a new probabilistic model that takes into account workers' attention to estimate the label quality. Expectation propagation is adopted for efficient  Bayesian inference of our model, and a generalized Expectation Maximization algorithm is derived to estimate both the ground truth of all tasks and  the label-quality of each individual crowd worker with  attention. In addition, the number of tasks best suited for a worker is  estimated according  to changes in attention.  Experiments against related methods on {three real-world and one semi-simulated datasets demonstrate that our method   quantifies the relationship between workers' attention and label-quality on the given tasks, and improves the aggregated labels.}
\end{abstract}

\section{Introduction}
Crowdsourcing is the process of collecting annotations of large-scale and complex data from online workers. It's an inexpensive mechanism that enables leveraging the power of crowds. Many crowdsourcing platforms (e.g.,  Amazon Mechanical Turk\footnote{https://www.mturk.com/}, CrowdFlower\footnote{https://www.figure-eight.com/}, and Baidu Test\footnote{http://test.baidu.com/crowdtest/}) have been developed and are widely-used. Tasks that are difficult for computers, but comparatively easy for humans, e.g. protein structure prediction \cite{cooper2010}, sequence alignment \cite{lakhani2013} and  sentiment analysis \cite{liu2012}, are successfully addressed with crowdsourcing.

In crowdsourcing, different workers provide  answers (annotations or labels) based on their domain of  expertise. Since workers usually have different skills with respect to a given task, answers from different workers  may be very noisy or, in some cases, inaccurate. Annotations provided by less competent workers are more error-prone. Furthermore, due to budget constraints, in practice, we may be able to collect only a  small number of answers, and sometimes no answer at all. In such cases, it is unreliable to infer the ground truth for each task using a simple algorithm like majority voting \cite{Sheng2008Get}.

Several approaches have been developed to derive high-quality answers. Ground truth inference algorithms  model the expertise of workers \cite{Whitehill2009Whose}, the biases of workers \cite{Zhuang2015Leveraging}, and the difficulty of tasks \cite{kurve2015multicategory}. Task assignment strategies assign tasks to workers with specific qualities \cite{fan2015icrowd}, or assign tasks to domain expert workers \cite{simpson2015bayesian}. \textit{All} these methods assume that the quality of the labels contributed by workers is fixed over time. However, in realistic crowdsourcing  scenarios, this is not true, {as the span of \textbf{\emph{attention}}\footnote{In this paper, the definition of attention is the ability to focus on finishing something. We study the process of the effect of attention on answering crowdsourcing tasks, rather than the process of attention being influenced by certain factors (e.g., fatigue, frustration, or some combination of other concepts). } }of a worker may change as he goes through the tasks to be labeled, and thus has impact on   the quality of the answers.
This problem of unstable answer quality should be investigated because
a) the answers provided by a focused worker are more reliable; b) when a worker's attention decreases, it's wise to stop assigning additional tasks to the same worker; and c) an incentive strategy can be adopted to stimulate  the worker for more reliable answers.
Therefore, we target on studying the \emph{attention-aware answers} of crowds.

In this paper, we propose a probabilistic  model (called A3C) for crowdsourcing attention-aware answers. A3C assumes that the label-quality of a worker varies as his attention changes during the labeling process. A3C adopts different  distributions to model the variation tendency of different types of workers (e.g., experts, normal workers, spammers), and leverages the features of each task to obtain reliable aggregated labels, based on the assumption that similar samples have similar labels. We use expectation propagation (EP) \cite{minka2001family} to perform an efficient approximate Bayesian inference of our probabilistic model. Based on the EP approximation inference, a generalized Expectation Maximization (GEM) algorithm is derived to estimate both the ground truth of all tasks and  the label-quality of each individual worker according to attention. The main contributions of our work are summarized as follows:
\begin{enumerate}
  \item To the best of our knowledge, A3C is the first approach that models the variation of a worker's label-quality from the perspective of his/her attention for a group of given tasks. A3C leverages the Gaussian,  Poisson, and Uniform distributions to model the variable attention of experts, normal workers, and spammers, respectively. The suitable number of tasks for experts and normal workers is also estimated.
  \item A3C uses a back-up mechanism to deal with noisy answers \cite{hernandez2011robust}. It enables robustness, especially when errors in labeling occur far from the decision boundaries. A3C further utilizes a generalized Expectation Maximization algorithm to estimate the ground truth and the  label-quality of each individual worker.
  \item Our extensive results validate the advantages of our proposed A3C approach over the competing solutions \cite{dawid1979maximum,zhang2016multi,Whitehill2009Whose,Zhang2017Consensus,ma2015faitcrowd} in aggregating answers. Attention models using the Poisson, Gaussian, and Uniform distributions are  explored. The results show that our method can explore the variation relationship between workers' attention and label-quality on the given tasks, especially for the \emph{normal workers}. We also study the reasons why the label-quality of experts and spammers is not influenced by variations in attention.
\end{enumerate}

The rest of the paper is organized as follows. We briefly review related work in Section \ref{sec:relwork},  and then elaborate on the proposed algorithm and its optimization in Section \ref{sec:ProbForm}. Section \ref{sec:exp} provides the experimental results and analysis, and Section \ref{sec:concl} discusses conclusions and ideas for future work.
\section{Related Work}
\label{sec:relwork}
Crowdsourcing utilizes the capabilities of the crowd to deal with computer-hard tasks. Due to the diverse backgrounds of   workers, answers may vary in quality. Several approaches  have been proposed  to achieve high quality answers in crowdsourcing. In the following, we review some representative solutions from the perspective of improving collected data quality and selecting workers.

\textbf{Improving data quality.} The most intuitive strategy to deal with the low quality of data
annotation is to improve the quality of the data itself
\cite{dawid1979maximum,Whitehill2009Whose,zhang2016multi,Zhang2017Consensus,chen2017learning,Tu2018multi}. The Dawid-Skene (DS) model \cite{dawid1979maximum} is a standard probabilistic model for label inference from multiple annotations using Expectation-Maximization (EM).
Whitehill \textit{et al.} \cite{Whitehill2009Whose} further modeled both  worker reliability and the difficulty of tasks using EM.  Zhang \textit{et al.} \cite{zhang2016multi} created probabilistic features for each task  and used a K-Means algorithm to cluster all tasks, where each cluster is mapped to a specific class label. However, they do not produce an estimate of workers' skills.   Zhang  \textit{et al.} \cite{Zhang2017Consensus} introduced an adaptive weighted majority voting (AWMV) approach. AWMV uses the frequency of positive labels in multiple noisy label sets of each task to estimate a bias rate, and then assigns weights, derived from the bias rate, to negative and positive labels. These methods tackle the problem of identifying low quality answers in different ways \cite{dawid1979maximum,Whitehill2009Whose,Zhang2017Consensus,chen2017learning,zhang2016multi,Tu2018multi}. However,  they all ignore the intrinsic \emph{features} of tasks, which can improve the quality of aggregated labels. {Ma \textit{et al.} \cite{ma2015faitcrowd} proposed a probabilistic model (FaitCrowd) to jointly model the process of generating the question's content and workers' answers to estimate both the topical expertise and correctness simultaneously.} {Zhang \textit{et al.} \cite{zhang2017label} proposed the BiLayer Clustering to cluster the conceptual-level and  physical-level features of tasks.
Atarashi \textit{et al.} \cite{atarashi2018semi} presented a generative deep learning model to leverage unlabeled data effectively by introducing latent features.}


\textbf{Selecting workers.} Another intuitive strategy is to identify workers that produce better results. Instead of waiting for tasks to be pulled by random workers, it may be more effective to proactively push tasks to the selected workers. If both the skills required by a task and those possessed by workers are defined, tasks can be  automatically assigned to matched workers. For example, Ipeirotis \textit{et al.} \cite{ipeirotis2010quality} used scalar scores to evaluate the quality of workers and to reject (or block) low-quality workers during the assignment of tasks. CrowdDQS selects workers with the high expected accuracy to complete the given task \cite{khan2017crowddqs}. Fan \textit{et al.} \cite{fan2015icrowd} proposed an adaptive crowdsourcing framework (iCrowd). iCrowd dynamically estimates the accuracy of a worker based on her/his performance on the completed tasks, and predicts which tasks are well fit for the worker. Kobren \cite{kobren2015getting} presented a model that predicts the `survival probability' of a worker at any given moment, and then leveraged this survival model to dynamically decide which tasks to assign.

Other approaches have been explored to obtain high-quality answers, such as giving incentives to workers, improving task design, and estimating the suitable number of workers for a task  \cite{drapeau2016microtalk}. Giving incentives acts on the motivation that pushes workers to perform well  \cite{ho2015incentivizing}. Methods on improving task design focus on refining the description or structure of a task, so that workers can easily understand the task \cite{rogstadius2011assessment}. Estimating the suitable number of workers for a task  entails quantifying the optimal number of workers needed for a given task, under budget constraints \cite{ho2012online}.

All the aforementioned solutions make the underlying assumption that the label-quality of a worker is \textbf{fixed} throughout the completion of an assignment. As previously argued, this assumption is violated in practice. {\cite{pattyn2008psychophysiological,goldberg2011boredom,dai2015and} show that long sequences of monotonous tasks incite boredom, or that underutilization is related to misdirection of attention resources or withdrawal. However, they did not concretely study the influence or change of the attention on the crowdsourcing tasks.} To achieve more realistic models, our A3C approach considers attention-aware answers of crowds. A3C uses different distributions to model the variation of workers' attention, and the impact that a changing attention has on the label-quality for a set of given tasks.  In addition, A3C leverages features of tasks to achieve more reliable aggregated labels. The proper number of tasks to be assigned to workers is also estimated.

\section{The Proposed Methodology}
\label{sec:ProbForm}
\subsection{Problem Definition}
In crowdsourcing, we are provided with a set $\mathcal{X}=\{\mathbf{x}_i\}_i^N$ of tasks, and a distinct label  set $\mathcal{C}=\{1,\cdots,C\}$. Each task has a $d$-dimensional feature vector $\mathbf{x}_i=[x_{i1}, x_{i2}, \cdots, x_{id} ]$ to represent its characteristics. Let $\mathbf{s}_i$ denote a latent random variable with a Gaussian process prior.
Suppose there are $W$ workers to annotate a group of $N$ tasks in any order, and the ordering of the labels of each task is assumed to be the same. Each worker $w$ either annotates a task once, or does not at all. This follows the rule of current crowdsourcing systems, where each task can be accepted and completed at most once by a specific worker. We denote the true label of $\mathbf{x}_i$ (usually unknown) as $y_i$. The label of $\mathbf{x}_i$ annotated by worker $w$ is $a_{iw}$, where $a_{iw}=c \in \{0\} \cup \mathcal{C}$. In particular,  $a_{iw}=0$ means that worker $w$ does not provide any answer for $\mathbf{x}_i$.  $\mathbf{a}_i= \{a_{iw}\}_{w=1}^W$ represents all the labels for $\mathbf{x}_i$ annotated by $W$ workers. Let $\mathbf{Y}=\{y_i,\cdots, y_N\}$, $\mathbf{S}=\{\mathbf{s}_i,\cdots, \mathbf{s}_N\}$, and $\mathbf{A}=\{\mathbf{a}_1,\cdots,\mathbf{a}_N\}$.

\begin{figure}
  \centering
   \includegraphics[height=3cm,width=8.5cm]{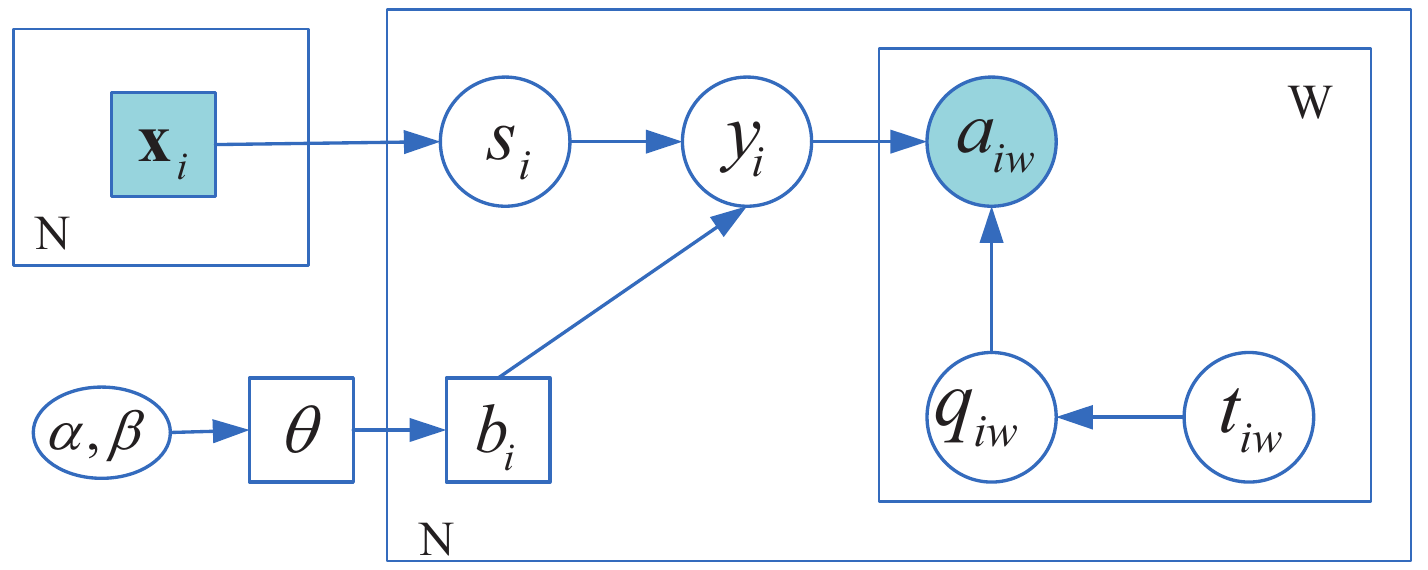}
   \vspace{-2em}
  \caption{The probabilistic graphical model of A3C. Circular nodes are random variables and square nodes are factor nodes. The shaded nodes represent observed values (worker annotations ($a_{iw}$) and task features ($\mathbf{x}_i$)).
  The model describes the process of generating an answer $a_{iw}$ for task $\mathbf{x}_i$ by worker $w$ with label-quality $q_{iw}$. $q_{iw}$ is influenced by the worker's attention $t_{iw}$.}\label{model}
  \vspace{-1em}
\end{figure}

Although the worker's attention changes over time, it is difficult to strictly mimic the time  factor. Given this, we use a surrogate, where the worker's attention changes with the number of completed tasks. This surrogate is reasonable, since the number of tasks completed by a worker is positively correlated with time. We propose a novel probabilistic generative model, illustrated in Figure \ref{model}. This model describes the process that generates noisy answers $\{a_{iw}\}_{i=1,w=1}^{N,W}$ of multiple workers of different label-quality $q_{iw}$. The worker's quality ($q_{iw}$) depends on the  attention $t_{iw}$, as the number of completed tasks increases.  Our goal is to study \textbf{the variational relationship between the worker's label-quality  and attention}, the number of tasks appropriate for each worker, and the true labels.

\subsection{Attention-Aware Probabilistic Model}
Given a group of tasks  $\mathcal{X}$, each worker $w$ (with label-quality $q_w$) independently completes a subset of tasks of $\mathcal{X}$, and provides the answers $a_{iw}$. The worker's attention changes as the number of completed tasks progresses, which in turn can influence $q_w$, and consequently $a_{iw}$. To approximate the above process, we define the conditional joint probability of our probabilistic model as follows:
\vspace{-0.1cm}
\begin{equation}\label{eq1}
\begin{aligned}
p(\mathbf{S}, \mathbf{Y}, \mathbf{A}, \mathbf{b}, \theta | \mathcal{X}, \alpha, \beta, t) \propto p(\theta) p(\mathbf{b}|\theta) p(\mathbf{S}|\mathcal{X}) \\
\cdot \prod_{i=1}^N\left\{p\left(y_{i} | \mathbf{s}_{i}, b_{i}\right) \prod_{w=1}^W p\left(a_{iw} | y_{i}, q_{iw}\right) p\left(q_{iw} | t_{iw}\right)\right\}
\end{aligned}\vspace{-0.1cm}
\end{equation}

where $p(\theta)$ denotes the probability with respect to the prior of the binary variables $\{b_i\}_{i=1}^N$. $p(\mathbf{b}|\theta)$ represents the probability of each $b_i$ under the condition $\theta$. $p(\mathbf{S}|\mathcal{X})$ calculates the probability of the random variable $\mathbf{s}_i$ with respect to each $\mathbf{x}_i$ . $p\left(y_{i} |\mathbf{s}_i, b_{i}\right)$ denotes the conditional probability of the true label. $p\left(a_{iw} | y_{i}, q_{iw}\right)$ denotes the probability that the worker with quality $q_{iw}$ and attention $t_{iw}$ gives the answer $a_{iw}$ for the task $\mathbf{x}_i$. $p\left(q_{iw} | t_{iw}\right)$  represents the conditional probability of the worker label-quality. {Note that the index $i$ is used to differentiate different tasks, and not to indicate the order according to which tasks are completed by the workers.}

In this model, $p(\mathbf{S}|\mathcal{X})$ is a Gaussian process prior with kernel tricks \cite{williams1998bayesian},
\vspace{-0.3cm}
\begin{equation}\label{eq2}
\small
p(\mathbf{S} | \mathcal{X})=\prod_{c=1}^{C} \mathcal{N}\left(\mathbf{S}_{c} | \mathbf{0}, \mathbf{K}_{c}\right)\vspace{-0.2cm}
\end{equation}

where $\mathbf{K}_c=[k(\mathbf{x}_i,\mathbf{x}_j)]_{i,j=1}^N$ is a kernel matrix defined over the tasks annotated with class label $c$. This treatment ensures that similar tasks have similar prediction scores. In theory, any valid kernel that measures the similarity among tasks (e.g., RBF and liner kernels) can be applied here.

{However, labeling errors in the context of Eq. (\ref{eq2}) are often contaminated with additive Gaussian noise, which can result in over-fitting problems when errors are actually observed far from the boundaries.  The following $p\left(y_{i} | \mathbf{s}_{i}, b_{i}\right)$ is a back-up mechanism \cite{hernandez2011robust} to deal with the issue,} and it's defined as follows:
\begin{equation}\label{eq3}
\small\vspace{-0.1cm}
p\left(y_{i} | \mathbf{s}_{i}, b_{i}\right)=\left[\prod_{c \neq y_{i}} \mathbb{I}\left(\mathbf{s}_{i, y_{i}}-\mathbf{s}_{i, c}\right)\right]^{1-b_{i}}\left[\frac{1}{C}\right]^{b_{i}}\vspace{-0.1cm}
\end{equation}
where $\mathbb{I}(x)=1$ if $x>0$ and  $\mathbb{I}(x)=0$, otherwise. A set of  binary latent variables $\mathbf{b}=\{b_i,\cdots, b_N\}$, one per task, is introduced to indicate whether $\mathbf{s}_{i,y_i} \geq \mathbf{s}_{i,c}$  for any $c= y_i$ $(b_i=1)$, otherwise $(b_i=0)$. We observe that the first term in Eq. (\ref{eq3}) directly depends on the accuracy of $\mathbf{s}_{i, y_{i}}$. In particular, it takes value 1 when the corresponding task is correctly classified, and 0 otherwise. Our model is robust when the observed data contain errors in labeling points which are far from the decision boundaries. This is because the likelihood function described in Eq. (\ref{eq3}) considers only the total number of prediction errors made by $\mathbf{s}_{i, y_{i}}$, rather than the distance between the wrongly predicted  tasks and the decision boundary.

$q_{iw} \in [0,1]$ represents the quality of the labels given by worker $w$ to task $i$. A larger $q_{iw}$ value leads to a higher probability that $a_{iw}$ will be consistent with the true label $y_i$. On the other hand, a smaller $q_{iw}$ results in a higher probability that worker $w$ will make mistakes. Thus, the conditional probability $p\left(a_{iw} | y_{i}, q_{iw}\right)$ is defined as follows:
\begin{equation}\label{eq4}
p\left(a_{i w} | y_{i}, q_{i w}\right)=q_{i w} \mathbb{I}\left(y_{i}=a_{i w}\right)+\left(1-q_{i w}\right) \mathbb{I}\left(y_{i} \neq a_{i w}\right)
\end{equation}
Alternative definitions of the quality of labels are also suitable for our model.

\begin{figure}
  \centering
   \includegraphics[scale=0.26]{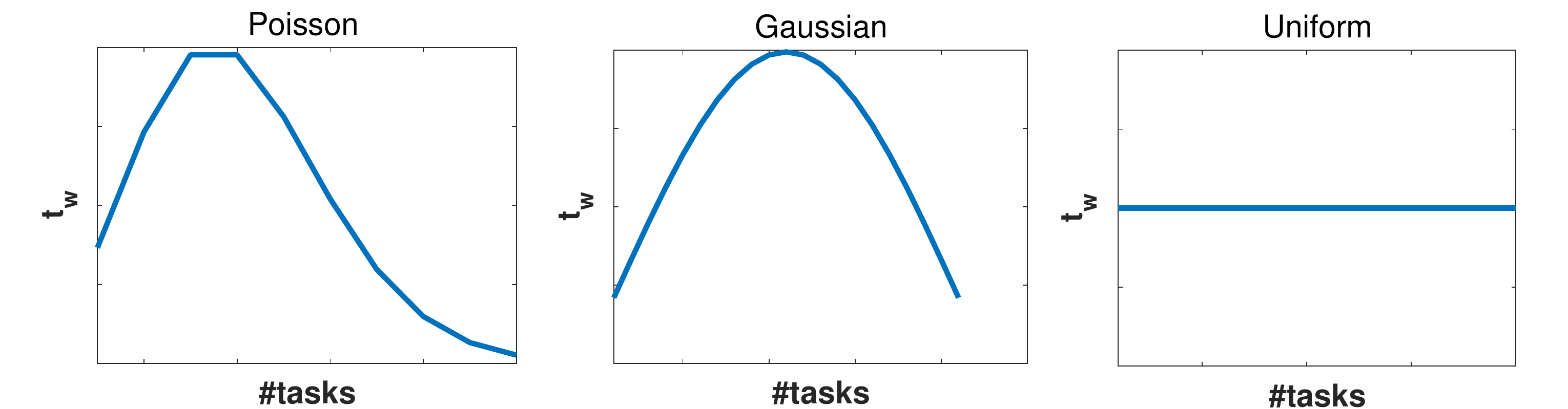}
     \vspace{-1em}
  \caption{Attention ($t_w$) vs. number of completed  tasks (\#tasks)  modeled under different distributions.} \label{fig:distributions} \vspace{-1em}
\end{figure}

In practice, when a worker performs a job on a group of given tasks, his/her attention varies as the completed tasks accumulate. It is usually difficult for the online worker to maintain a high attention level to complete all the given tasks. Suppose the worker's attention gradually increases at the start, and when a certain amount of tasks is completed, the worker's attention decreases. {To model this process, we use a Poisson distribution\footnote{The standard formula for the Poisson distribution is $P(k,\lambda)=\frac{\lambda^k}{k!}e^{-\lambda}$. Here, we use the Stirling formula ($k! \approx \sqrt{2\pi}k^{k+\frac{1}{2}}e^{-k}$) \cite{robbins1955remark} to approximate the probability of Poisson distribution.} (shown in Figure \ref{fig:distributions} (left)) to describe the variation of the worker's attention as follows:}
\begin{equation}\label{eq5}
t_{iw}\sim Poi(\frac{N_w}{{\lambda}_{w}}) \approx (2 \pi i)^{-1 / 2} e^{-\frac{N_w}{{\lambda}_{w}}}\left(\frac{N_w e}{{\lambda}_{w}i}\right)^{i}
\end{equation}
where $N_w$ is the number of tasks given to worker $w$, and $N_w/ \lambda_{w}$ is the parameter that captures the variation of a worker's attention. The expectation and variance of the Poisson distribution are both $N_w/ \lambda_{w}$.  ${\lambda}_{w}$ also indicates the suitable number of tasks for the $w$-th worker. If the number of annotated tasks is larger than ${\lambda}_{w}$, the attention will decrease. This phenomenon will be explored in the experiments. Other distributions can be used to describe a varying attention. We  investigate the uniform  and Gaussian distributions for different types of workers. The variation of attention under different distributions is shown in Figure \ref{fig:distributions}.

It is recognized that a worker is more likely to make mistakes  when his level of attention is low. Vice versa, a worker with high attention level tends to give high-quality answers. In this paper, for simplicity, $p\left(q_{iw} | t_{iw}\right)$ is defined as:
\vspace{-0.2cm}
\begin{equation}\label{eq6}\vspace{-0.2cm}
p\left(q_{i w} | t_{i w}\right) \propto t_{i w}
\end{equation}

We do not have any preference for a particular task to be considered  as an outlier. Thus,  $p(\mathbf{b}|\theta)$ is set to a prior factorizing multivariate Bernoulli distribution as follows:
\vspace{-0.2cm}
\begin{equation}\label{eq7}
\small
p(\mathbf{b} | \theta)=\prod_{i=1}^{N} \theta^{b_{i}}(1-\theta)^{1-b_{i}}
\vspace{-0.5em}
\end{equation}
where $\theta$ is the prior fraction of training instances expected to be outliers.  The prior $\theta$ is unknown and thus is set to be a conjugate beta distribution as:
\vspace{-0.2cm}
\begin{equation}\label{eq8}
\small
p(\theta)=\operatorname{Beta}(\theta | \alpha, \beta)=\frac{\theta^{\alpha-1}(1-\theta)^{\beta-1}}{\operatorname{Beta}(\alpha, \beta)}
\end{equation}
where $\alpha$ and $\beta$ are free hyper-parameters. The input values of $\alpha$ and $\beta$  do not have a large influence on the final model, provided that  most of the tasks are correctly aggregated. In addition, their values are kept small so that Eq. (\ref{eq8}) is not too constraining. In this paper, the values of $\alpha$ and $\beta$ estimated by a generalized Expectation Maximization (see Section \ref{gem}) are similar, and they do not influence the performance of our method.  We finally set $\alpha=2$ and $\beta=9$.

\subsection{Learning with Generalized Expectation Maximization}
\label{gem}
In order to evaluate the quality and attention of multiple workers, we need to estimate the parameter $\Theta=\{\alpha,\beta,\{\lambda_w\}_{w=1}^W, \{q_{iw}\}_{i=1,w=1}^{N,W}\}$. Thus, we use a generalized Expectation Maximization algorithm (GEM) with the lower bound of the log likelihood defined as follows:
\begin{equation}\label{eq17}
\small
\begin{aligned}
&\log p\left(\mathbf{A}, \mathbf{S} | \mathcal{X}, \Theta\right)
\\
&\quad \geq \sum_{b_{i}} \int_{\mathbf{S}} Q\left(\mathbf{S}\right) \log \frac{p(\mathbf{b} | \theta) p(\theta) p\left(\mathbf{T}, \mathbf{S} | \mathcal{X}, \Theta\right)}{Q\left(\mathbf{S}\right)} \\
&\!=\!\!\mathrm{C}\!\!+\!\!\sum_{b_{i}} \sum_{i=1}^{N} \int_{\mathbf{s}_{i}} q\left(\mathbf{s}_{i}\right) \log \left\{p\left(b_{i} | \theta\right) p(\theta) p\left(\mathbf{a}_{\mathbf{i}} | \mathbf{s}_{i}, b_{i}, q_{iw}, t_{iw}\right)\right\} d \mathbf{s}_{i}\end{aligned}
\end{equation}
The GEM works as follows.

\textbf{E-Step:} Given the current parameter $\Theta_p$, the features $\mathcal{X} $ and the observed answers $\mathbf{A}$, we conduct the Expectation Propagation inference to obtain an approximate inference of $Q\left(\mathbf{S}\right) \sim p\left(\mathbf{S} | \mathcal{X}, \mathbf{A}, \Theta_{p}\right)$.

\textbf{Generalized M-Step:} To achieve a new $\Theta$, we maximize the lower bound of $\log p(\mathbf{A},\mathbf{S}|\mathcal{X},\Theta)$, and then replace $\Theta_p$. Since the  closed-form solution of $\Theta$ is intractable,  the L-BFGS-B algorithm \cite{zhu1997algorithm}  is used to find a numerical estimation that maximizes the lower bound in Eq. (\ref{eq17}) by gradient ascent, which is guaranteed to obtain a local optimal solution.
\section{Experiments}
\label{sec:exp}
\subsection{Experimental Setup}
\label{setup}
{\textbf{Datasets:}
 We use four (three real and one semi-synthetic) datasets for the experiments.  \textbf{Dog} \cite{liu2019interactive} contains 800 images of 4 breeds of dogs from ImageNet \cite{deng2009imagenet}. The images are annotated by 52 workers. \textbf{Bird} \cite{liu2019interactive} contains 2000 instances  with 10 category labels from ImageNet \cite{deng2009imagenet}. The annotations are collected from AMT and are made by 65 workers. \textbf{Game} \cite{aydin2017crowdsourced} is collected from a crowdsourcing platform via an Android App based on the TV game show. It contains 2,103 questions, 37,029 workers, 214,849 answers, and 12,995 unique words. {We select a subset of workers (200) and their answers by Gibbs sampling. }\textbf{News} \cite{lang1995newsweeder} contains documents of four topics from the 20NewsGroup dataset. Note that, although the global true labels are known, we only use them when we calculate the evaluation metric.

Based on the classification in \cite{vuurens2011much,kazai2011worker}, we simulated the crowdsourced labels for each instance in the \textbf{News} dataset with the protocol used in \cite{hung2013evaluation}.
The Dog, Bird, and News datasets only contain worker annotations and tasks without features. To measure the similarity between tasks, we extracted a feature vector for each task. For images, we extracted feature vectors using a deep convolutional neural network, namely VGG-NET \cite{simonyan2014very}. We used the output of the last fully connected layer as the feature vector of the image. The number of features is 1000.  For a document, we extracted its feature vector using TF-IDF. The dot-product kernel defined in Eq. (\ref{eq2})  is used to measure the similarity between tasks.}

\textbf{Evaluation metric:} We compare our A3C with the baseline methods in terms  of \textbf{Accuracy}.

\textbf{Comparing methods:} We compare  A3C against five methods, MV, DS\cite{dawid1979maximum}, GLAD\cite{Whitehill2009Whose}, FaitCrowd \cite{ma2015faitcrowd}, GTIC\cite{zhang2016multi}, and AWMV\cite{Zhang2017Consensus}. All were reviewed in the Related Work Section.
All the comparing algorithms run on the whole dataset to calculate the class label of all tasks. The evaluation metric is then computed against the ground truth.  The initial values of the parameters for each algorithm are set as recommended by the authors. The DS and GLAD (EM-based) algorithms stop at convergence. To avoid oscillations of the target function around a local optimum, we also set a maximum number of iterations (50). For GTIC, we set $K=C$, where $K$ is the number of clusters for $k$-means.  We set $\alpha=2, \beta=9$ for A3C.

\subsection{Comparison against state-of-the-art methods}
{It is difficult to extend the comparing  methods to the attention scenario. Therefore, we introduce A3C(nA), a variant of A3C, which does not consider attention. A3C(nA) sets $p(q_{iw}|t_{iw})=1$. We compare A3C(nA) against the state-of-the-art methods under the no-attention scenario, to  show the performance of A3C(nA) and the comparing methods on class label aggregation.

The results are given in Table \ref{Accuracy}. FaitCrowd is designed for text data, and therefore its accuracy is given for Game and News only. We have the following observations: 1) A3C(nA) achieves the highest Accuracy on all four datasets.  Both A3C(nA) and FaitCrowd outperform the other methods on the Game and News datasets. This is because A3C(nA) and FaitCrowd take into account the intrinsic features of the tasks, while the others do not. A3C(nA) assumes that similar tasks have similar class labels, and this largely improves label aggregation. The back-up mechanism used in A3C(nA) is robust to noise and can further improve the performance. This is why A3C(nA) achieves better results than FaitCrowd;  2) GTIC often outperforms  MV, DS, and AWMV, because it has the ability of grouping examples with similar feature patterns into clusters. In addition, it is generally capable of assigning the most appropriate class label to each task; 3) DS almost always outperforms GLAD; this phenomenon can be explained by the fact that DS models the entire confusion matrices of all workers, and the probabilities of all classes; 4) MV always achieves the worst results, since it does not account for the features of tasks, and connections between answers.  From these observations,  we can draw the conclusion that A3C(nA) (without considering workers' attention) can  effectively aggregate the workers' answers by utilizing the back-up mechanism and the intrinsic features of tasks.}
\begin{table}[h!tbp]
\center
\vspace{-0.5cm}
  \caption{Accuracy of A3C(nA) and the comparing methods on the four original datasets on aggregating class labels. The best results are in \textbf{boldface}. All the variances are typically less than 1\% for 10 runs and omitted here. A3C(nA) does not model attention variance.}
  \vspace{-0.2cm}
  \resizebox{0.5\textwidth}{!}{
  \begin{tabular}{lccccccccc}
    \toprule
    \bfseries Datasets &
    \bfseries MV &
    \bfseries GLAD &
    \bfseries DS &
    \bfseries AWMV &
    \bfseries GTIC &
    \bfseries FaitCrowd &
    \bfseries A3C(nA) &\\
    \midrule
    Dog & 0.826 & 0.831 &0.842 &0.861 &0.852 &- &\textbf{0.875}\\
    Bird & 0.813  & 0.818  & 0.822  &0.81 & 0.815 &- & \textbf{0.867}\\
    Game& 0.815  & 0.845  & 0.862  &0.884 & 0.895 &0.912 & \textbf{0.920}\\
    News& 0.672  & 0.744  & 0.759  &0.784  & 0.821 &0.849 &\textbf{0.861}\\
    \bottomrule
    \end{tabular}}%
\label{Accuracy}
\vspace{-0.3cm}
\end{table}%

\begin{figure}
  \centering\vspace{-0.5cm}
   \includegraphics[height=3.8cm,width=8.6cm]{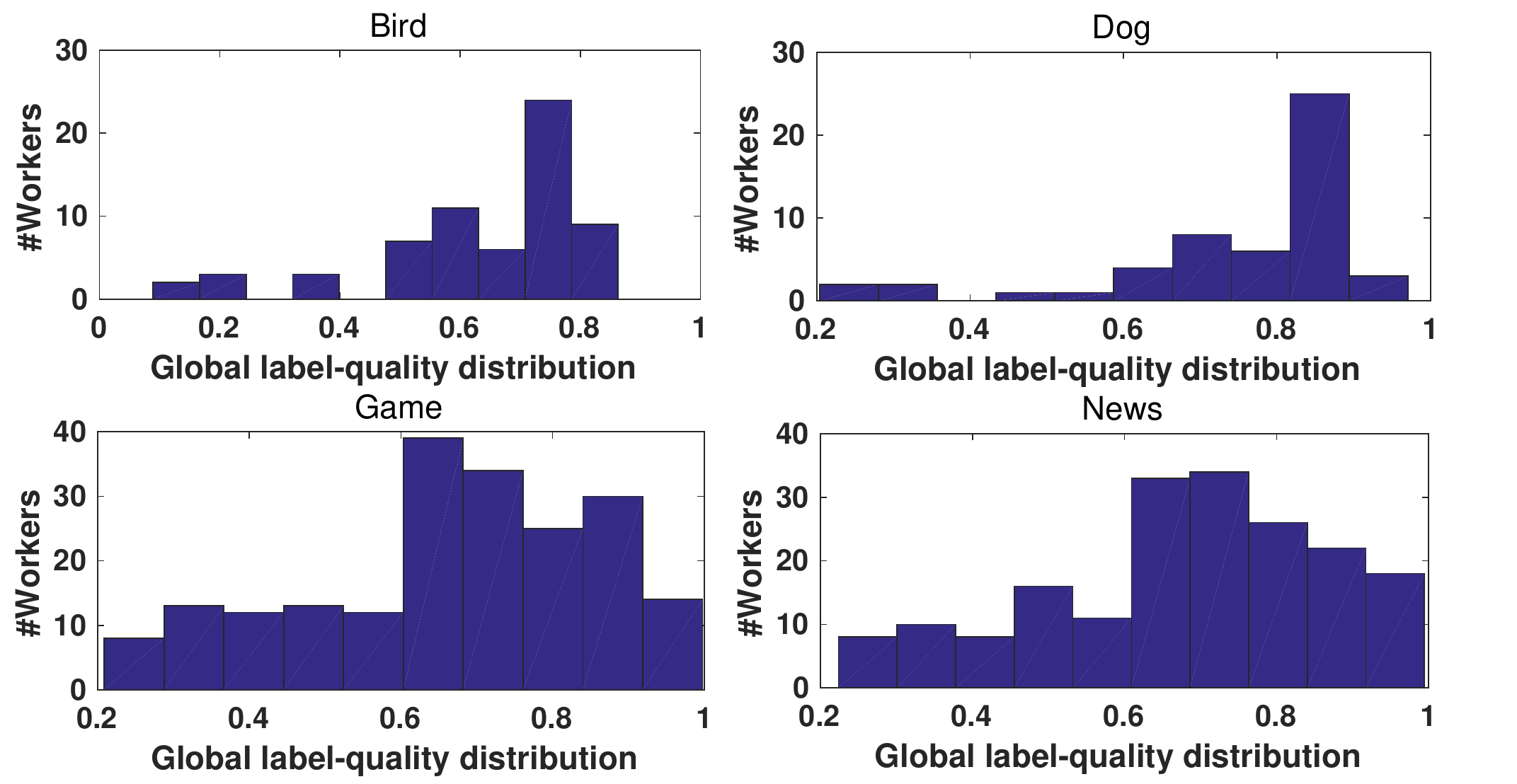}
  \caption{Global label-quality distribution of all workers on different datasets. }\vspace{-0.7cm}\label{workerquality}
\end{figure}

Figure \ref{workerquality} shows the global label-quality  distribution of all workers on the four datasets, where the  label-quality of each worker is computed by A3C(nA), using a generalized Expectation Maximization algorithm as in Eq. (\ref{eq17}). The global label-quality of each worker is defined as the probability that he correctly annotates the tasks.  A3C(nA) can effectively estimate the global label-quality of each worker on different datasets. {From the whole distribution of different datasets,} we find that the proportion of the workers whose global label-quality $q_w \geq 0.6$ within 60\%$\sim$70\%, and the proportion of the workers whose global label-quality $q_w < 0.4$ is about 12\%. By referring to  the statistical results in \cite{vuurens2011much,kazai2011worker} and the evaluation in \cite{hung2013evaluation}, the proportion of $q_w \geq 0.6(q_w < 0.4)$ is about 60\% (9\%), which are close to the specified portions of experts, normal workers and spammers. Taking the Bird dataset as an example, the number of workers whose $q_w \geq 0.6$ is about 42 (64.6\%).  In addition, A3C(nA) identifies the low-quality workers (even Spammers), whose label-quality is  smaller than 0.5. For the real-world datasets, we do not know the actual value of the worker's label-quality. Therefore, we use the known  $\{q_w\}_{w=1}^W$ values on the News dataset, to further verify the reliability of global label-quality $q_w$ of workers estimated by A3C(nA). We find that the difference between the estimated and true values is less than 0.02. Thus, we can draw the conclusion that A3C(nA) can effectively and accurately estimate the global  label-quality of  workers.

\subsection{Attention-aware Answers with Different Attention Distributions}\label{attention-distribution}

\subsubsection{Poisson Distribution.} To understand the behavior of A3C, we simulate attention-aware answers on all the datasets. On the Bird dataset, the global label-quality estimated by A3C(nA) can be viewed as an average on the completed tasks. We let the attention of worker $w$ follow a Poisson distribution. The worker's attention and the label-quality are positively correlated. Thus, the label-quality of worker $w$ follows a Poisson distribution, and the average is $q_w$. When given a task $\mathbf{x}_i$, the worker will re-annotate it according to the label-quality $q_{iw}$. Finally, we can obtain a new semi-synthetic Bird dataset with worker attention, called Bird(P). For News dataset, instead of a fixed label-quality, we adopt a variational quality of workers following a Poisson distribution. All the workers re-answer the tasks using \cite{hung2013evaluation}.

\begin{figure*}[h!tbp]
  \centering\vspace{-2cm}
   \includegraphics[height=5.2cm,width=15cm]{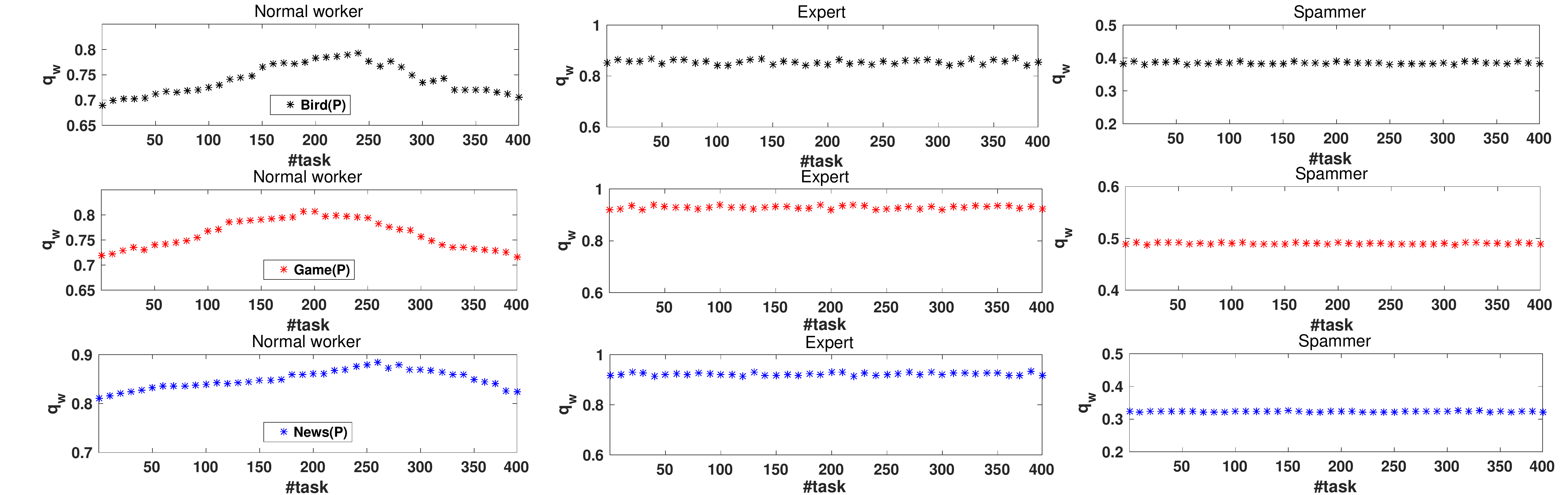}
   \vspace{-0.5cm}
  \caption{Label-quality $q_{iw}$ vs. number of accomplished tasks with respect to different types of workers (Normal, Expert, and Spammer). Workers' attention follows a Poisson distribution (Figure \ref{fig:distributions}, left). }\label{possion-quality}
\end{figure*}

For further analysis, according to \cite{hung2013evaluation,kazai2011worker}, we select an expert ($q_{iw}\geq0.9$), a normal worker ($q_{iw}\in [0.6,0.9)$) and a spammer ($q_{iw}<0.5$). Figure \ref{possion-quality} reports the partial results of the label-quality with respect to the three workers on each dataset. From Figure \ref{possion-quality}, we have the following observations: 1) For normal workers, as the number of accomplished tasks increases, the label-quality on different tasks ($q_{iw}$) first goes up, and then decreases. A3C assumes that the label-quality ($q_{iw}$) and the attention ($t_{iw}$) are positively correlated; that is, the workers' attention can influence the label-quality of the worker on a given task. The higher the attention is, the higher the label-quality is. In addition, the extent of the decline is relatively large, i.e. when attention decreases, the worker is more likely to make  mistakes. These phenomena are consistent with actual crowdsourcing: a normal worker has an initial warm-up period, until he/she reaches complete focus on the task. With time, the attention will decrease due to fatigue or other factors. 2) For experts, whose global label-quality is very high, the variation of  attention for a given task is small. This is because experts are generally more careful on providing answers, or can easily annotate the tasks with correct answers, due to their expertise. 3) Spammers simply provide random or uniform answers; as such, attention has close to zero influence on their label-quality.

\subsubsection{Gaussian distribution.}
\begin{figure*}[t]
  \centering
   \includegraphics[height=5.2cm,width=15cm]{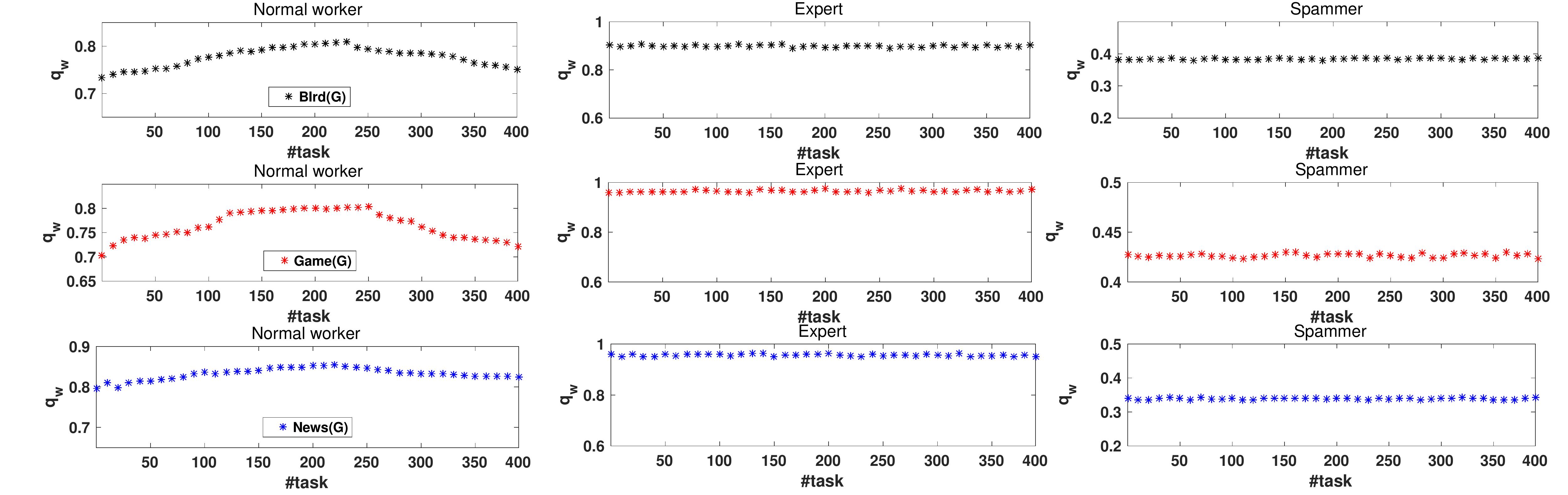}\vspace{-0.5cm}
  \caption{Label-quality $q_{iw}$ vs. number of accomplished tasks with respect to different types of workers (Normal, Expert, and Spammer). Workers' attention follows a Gaussian distribution (Figure \ref{fig:distributions}, middle). }  \vspace{-1.5em}\label{gaussian-quality}
\end{figure*}

We conduct additional experiments to investigate the impact of different attention distributions. In this case, we let the worker's attention $t_w$ follow a Gaussian distribution. As for the Gaussian distribution, we obtain three new datasets, called Bird(G), News(G), and Game(G). To explore the crowd answers  and the relationship between the  worker's label-quality and attention, we substitute $t_{iw}$ in Eq. (\ref{eq5}) with a Gaussian distribution with  mean $\frac{N_w}{\mu_w}$ and variance  $\sigma_{w}^2$. That is, $t_{iw} \sim N(\frac{N_w}{\mu_w}, \sigma_{w}^2)$. The same Expectation Propagation and a Generalized Expectation Maximization are used to estimate both the ground truth of all tasks and the label-quality of each individual worker. We select the same expert, normal worker and spammer as in Figure \ref{possion-quality}, and report the results under the Gaussian distribution in Figure \ref{gaussian-quality}.

From Figure \ref{gaussian-quality}, we observe the following: 1) For the normal worker, the label-quality increases at the beginning and decreases as the completed tasks increase. This observation confirms that workers' attention can influence the label-quality on a given task; 2) For the same reasons discussed under the Poisson distribution, the label-quality of the expert and the spammer workers is  nearly invariable to changes in attention.

Comparing Figures \ref{possion-quality} and \ref{gaussian-quality}, we can find that: 1) The Poisson  and the Gaussian distributions can capture the relationship between workers' attention and label-quality on a given task. 2) Normal workers'  attention is usually relatively low at the beginning, and  goes up  as the number of assigned  tasks increases, but decreases when the completed tasks exceed a certain number. 3) For the normal worker, the extent of the decline  is different for different distributions.  The Poisson distribution declines faster than the Gaussian. These phenomena are consistent with the nature of the two distributions (Figure \ref{fig:distributions}). 4) The label-quality of the expert and the spammer workers is not affected by attention. Overall, our method can effectively explore the relationship between workers' attention and the label-quality on  given tasks, especially for normal workers.

\subsubsection{Uniform distribution.}
\begin{figure}
  \centering
   \includegraphics[scale=0.28]{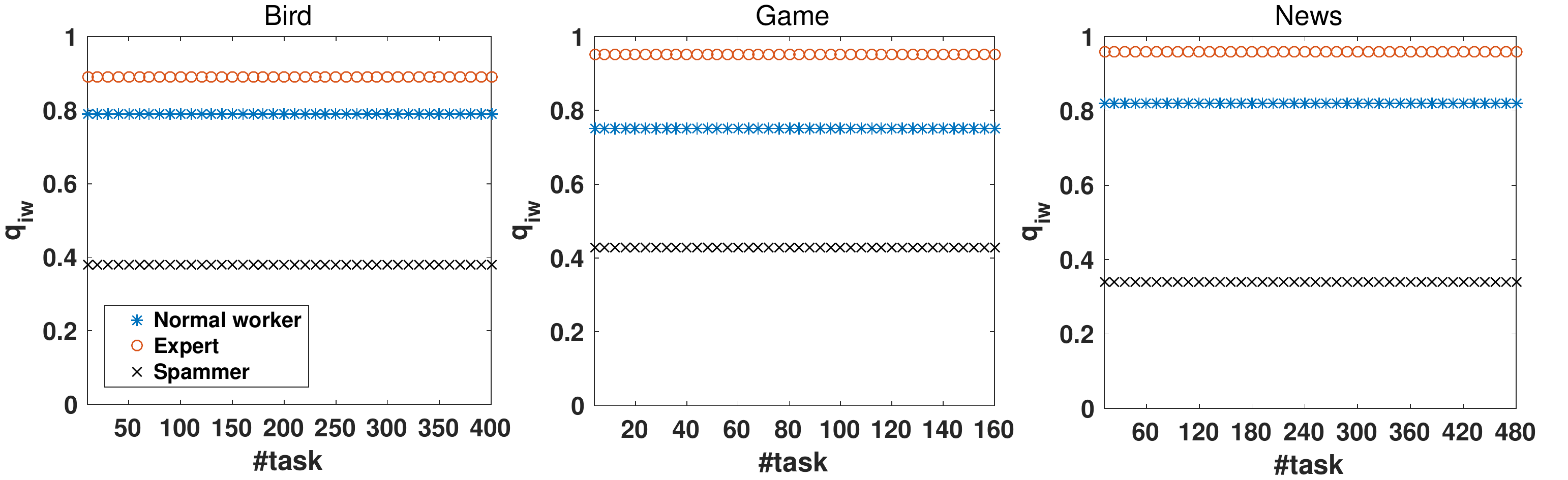}\vspace{-0.4cm}
  \caption{Label-quality $q_{iw}$ vs. the number of accomplished tasks with respect to different types of workers (Normal, Expert, and Spammer). Workers' attention follows a uniform  distribution (Figure \ref{fig:distributions}, right). }\vspace{-0.5cm}\label{uniform-quality}
\end{figure}

Attention under the uniform distribution is also studied. In this case, the attention on all given tasks does not change. Thus, the label-quality will not change.  That is, $p(q_{iw}|t_{iw})=1$ in Eq. (\ref{eq6}). We report the label-quality of the same normal worker, expert, and spammer in Figure  \ref{uniform-quality}.
As expected, each type of workers has the same label-quality on all given tasks. We leave the study of complex distribution mixtures to future investigation. {In addition, workers answer tasks in any order. A worker may finish one task ahead of others. Therefore, we select three workers, instead of computing the average of workers' performance in each category to show the relationship between the label quality and attention.}

{To explore the effect of attention on aggregating truths, we compare A3C against with A3C(nA) on the semi-synthetic datasets with Poisson distribution.  Table \ref{Accuracy-A3c-A3c(nA)} shows that A3C outperforms A3C(nA) on aggregating true labels. This is because A3C considers the impact of workers' attention on their label-quality while A3C(nA) does not.  A3C can utilize the high-quality answers and reduce the impact of low-quality answers annotated with lower attention. Compared with Table \ref{Accuracy}, the Accuracy of A3C(nA) gets a bit lower, that is because some answers are changed according to the label-quality when generating the semi-synthetic data. In addition, we test on the semi-synthetic datasets with Gaussian distribution, we get the similar results and conclusions as with Poisson distribution.}
\begin{table}[h!tbp]
\scriptsize
  \caption{Accuracy of A3C and A3C(nA) on aggregating class labels with attention in Poisson distribution (P) and Gaussian distribution (G).}
  \begin{tabular}{lcccccc}
    \toprule
    \bfseries Datasets &
    \bfseries Dog(P)/(G)&
    \bfseries Bird(P)/(G)&
    \bfseries Game(P)/(G)&
    \bfseries News(P)/(G) &\\
    \midrule
    A3C & \textbf{0.875/0.863} & \textbf{0.876/0.872} &\textbf{0.932/0.944} &\textbf{0.861/0.866} \\
    A3C(nA) & 0.860/0.852& 0.851/0.860  & 0.910/0.918 &0.855/0.841 \\
    \bottomrule
    \end{tabular}%
\label{Accuracy-A3c-A3c(nA)}\vspace{-0.5cm}
\end{table}%

\subsection{Analysis on the Estimated Number of Tasks}
When studying the variation of workers' attention, it's important to explore  how many tasks are suitable for the worker under high attention conditions.  $\boldsymbol{\lambda}$ in Eq. (\ref{eq5}) indicates the suitable number of tasks for a worker. From the previous analysis, we have seen that both  experts and spammers are  not influenced by attention. Thus, as the evaluation done in \cite{hung2013evaluation}, we report the results of the normal workers, whose maximal label-quality  is $\max\{q_{iw}\} \in [0.6, 0.9)$, on  Bird(P), Game(P), and News(P). We sort the selected workers in ascending order according to the label-quality, and report the values of $\boldsymbol{\lambda}$ in Figure \ref{lambda}. {We assume that the task requester does not assign new tasks to a worker when the attention of the worker starts to fall. We realize that workers can still complete tasks with relatively high $q_{iw}$ when the attention starts to fall and is still within a certain range. We consider ${\lambda}_{w}$ a reasonable value because i) the exact number of tasks suitable for a worker is hard to estimate due to the complexity of real-world crowdsourcing; and ii) if the worker's attention drops too much, $t_{iw}$ can still be stimulated by other factors.

Figure \ref{lambda} shows the following. 1) A3C can compute customized   ${\lambda}_{w}$ values for different workers. 2) Workers with low quality have small ${\lambda}_{w}$ values, which suggests that workers with  low label-quality should not be assigned too many tasks. Different workers are suitable for different numbers of given tasks, which is consistent with the realistic scenario. 3) As expected, workers with higher label-quality can complete more tasks. By studying the variation of a worker's attention, we can estimate ${\lambda}_{w}$, which can be used to collect good answers with confidence. In addition, the value of  ${\lambda}_{w}$ can be stored and leveraged for a  worker $w$ for future crowdsourcing tasks.
\begin{figure}
  \centering \vspace{-1cm}
   \includegraphics[scale=0.28]{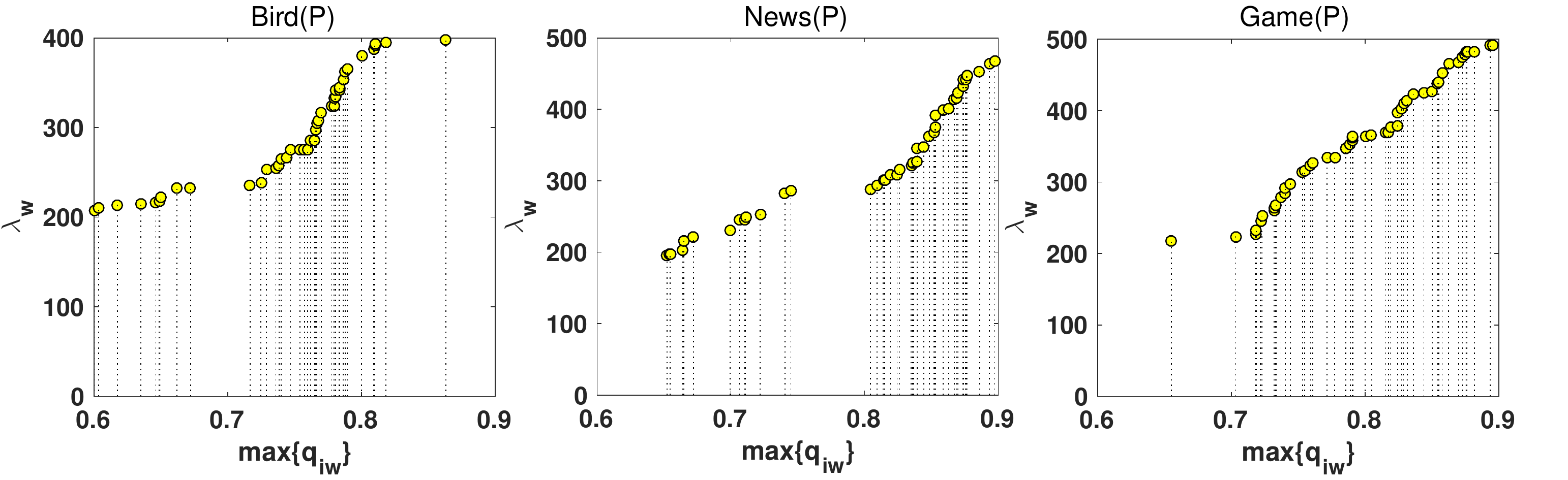}\vspace{-0.4cm}
  \caption{ $\lambda_w$ vs. the maximal label-quality ($\max\{q_{iw}\}$) in ascending order on different datasets with the Poisson distribution.}\vspace{-0.6cm}\label{lambda}
\end{figure}

When attention follows the Gaussian distribution, the value of $\mu_{w}$ reflects the  suitable number of tasks for a worker.  Following the same process as in Figure \ref{lambda}, we report the results of normal workers, whose maximal label-quality is $\max\{q_{iw}\} \in [0.6, 0.9)$, on  Bird(G), News(G), and Game(G). The results are shown in Figure \ref{mu}. We can see that A3C can estimate individual values for $\mu_{w}$. Again, workers with low label-quality  should be assigned fewer tasks, while  workers with high label-quality can complete more tasks. These observations are similar to those of Fig. \ref{lambda}. The difference between Fig. \ref{lambda} and Fig. \ref{mu} is that the appropriately suitable number of tasks for a worker estimated by A3C is different, but adjacent. In this paper, we adopt the Poisson distribution for two reasons: 1) it contains only one parameter and can be more easily estimated; 2) the tendency of studied $q_{iw}$ is relatively more suitable for the normal workers.
\begin{figure}
  \centering\vspace{-1cm}
   \includegraphics[scale=0.28]{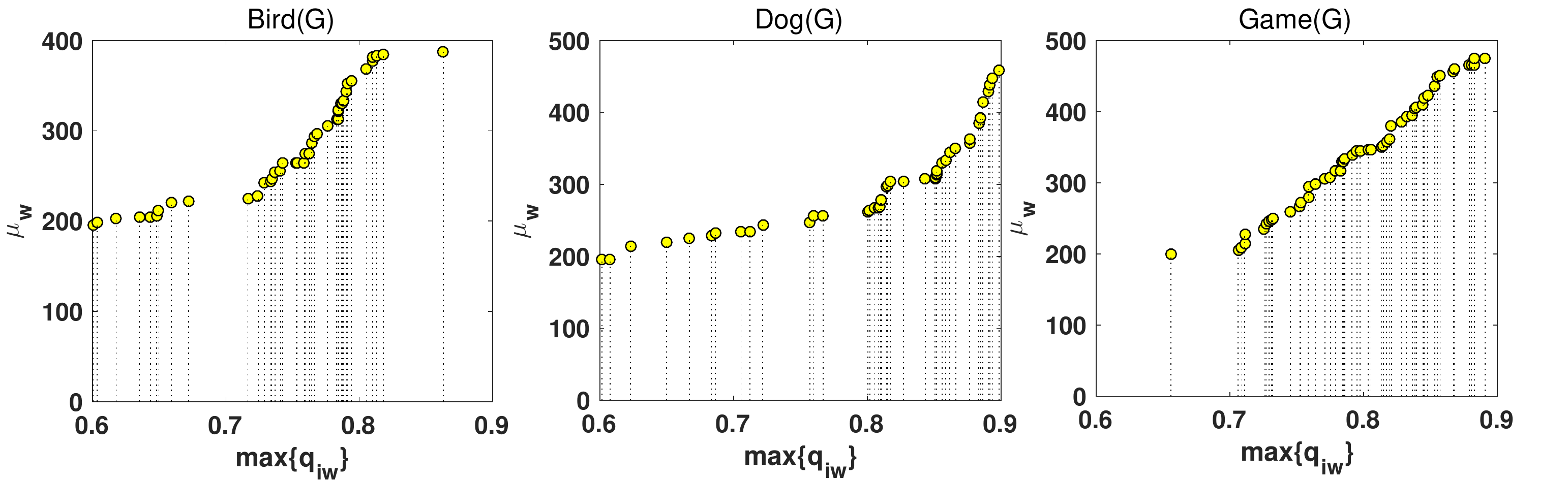}\vspace{-0.4cm}
  \caption{ $\mu_w$ vs. the maximal label-quality ($\max\{q_{iw}\}$) in ascending order on different datasets with the Gaussian distribution.}\vspace{-0.6cm}\label{mu}
\end{figure}

{From the results and analysis above, we can conclude that A3C can learn the attention, labeling-quality, and approximately suitable number of tasks for a worker. In this way, a task requester can obtain a relatively high-quality answers by rejecting spammers, and reasonably assign tasks  within the budget. }

\section{Conclusion}\label{sec:concl}
In this paper, we explore attention-aware answers of crowds, which is a novel, realistic, but unexplored scenario of crowdsourcing.  We develop a probabilistic model approach called A3C to tackle the problem. A3C assumes that the label-quality of a worker changes over time as his attention level also changes. We adopt different  distributions to model the variation trend of different types of workers (e.g., experts, normal workers, and spammers). We perform extensive experiments on {three real-world and one semi-synthetic datasets. The results show that A3C outperforms other related methods in aggregating labels. In addition, A3C can effectively explore the  relationship between the worker's quality and attention, and accurately estimate the label-quality and the suitable number of tasks of each worker.}
In the future, we will incorporate the worker's expertise into attention-aware crowdsourcing, and investigate online crowdsourcing with different attention distributions.

\bibliographystyle{IEEEtran}
\bibliography{A3C_Bib}

\end{document}